\documentclass[sigconf,nonacm]{acmart}

\IfFileExists{libertine.sty}{}{\usepackage{mathptmx}}
\usepackage{tikz}
\usepackage{balance}
\usetikzlibrary{positioning}

\setcopyright{none}

\begin{document}

\title{An Error Budget for Shuffled Multimodal Windows in the 3rd WEAR Challenge: Stacking, Count Constraints and Seam Matching}

\author{Soheil Kooklan}
\orcid{0009-0003-5035-7833}
\affiliation{%
  \institution{Independent Researcher}
  \country{}}
\email{soheilkooklan@gmail.com}

\begin{abstract}
The 3rd WEAR Dataset Challenge asks for an activity label for every one-second window of four unseen participants. Each window has 15 VideoMAEv2 frame embeddings and the accelerometer signal of one random limb, and the windows are shuffled. This report is an empirical analysis of one system built after the challenge deadline. Window classifiers are stacked with features averaged over per-participant graphs of video neighbours and scored successor candidates, and a mean-field style decoder with a soft count constraint per activity labels all windows of a participant together. The setting is offline and transductive, and the count constraint relies on each activity being performed once. In a participant-wise simulation with participant-level bootstrap intervals, stacking added 0.051 macro F1 over stage-1 fusion and the count constraint 0.011 (95\% interval 0.006 to 0.017). A quadratic seam distance ranks the true same-limb successor first among about 720 candidates in 83.9\% of cases, but adding it to the link model changed macro F1 by $-0.0001$ ($-0.0006$ to $0.0005$). With the decoder settings and the mixing weight chosen inside the folds and the mislabelled spans left out of training, the system scores 0.936 in simulation (0.940 for participants who performed each activity once); the submitted configuration scores 0.938 (0.943). Per-participant scores range from 0.80 to 0.97. These results come from a non-nested stacking protocol that uses unlabelled features transductively, so they are not a fully inductive estimate. An oracle that averages the final unary scores over the true segments reaches 0.973, and 79\% of the remaining errors involve the null class. On the private leaderboard, which no choice used, the selected Kaggle submission scored 0.9231 (public 0.9254, with bounds partly tuned there), and a later one with bounds chosen in simulation 0.9247 (public 0.9253).
\end{abstract}

\begin{CCSXML}
<ccs2012>
<concept>
<concept_id>10003120.10003138</concept_id>
<concept_desc>Human-centered computing~Ubiquitous and mobile computing</concept_desc>
<concept_significance>500</concept_significance>
</concept>
<concept>
<concept_id>10010147.10010257</concept_id>
<concept_desc>Computing methodologies~Machine learning</concept_desc>
<concept_significance>300</concept_significance>
</concept>
</ccs2012>
\end{CCSXML}
\ccsdesc[500]{Human-centered computing~Ubiquitous and mobile computing}
\ccsdesc[300]{Computing methodologies~Machine learning}

\keywords{human activity recognition, inertial sensors, egocentric video, multimodal fusion, graph decoding, WEAR dataset}

\maketitle

\section{Introduction}

WEAR is an outdoor workout dataset recorded with four smartwatches on the wrists and ankles, whose inertial measurement units (IMUs) give 3D acceleration at 50\,Hz, and a head-mounted camera~\cite{bock2024wear}. The paper describes the recording plan as roughly 90 seconds per activity with breaks allowed~\cite{bock2024wear}, and the project page as at least 90 seconds per activity~\cite{wearweb}. The 3rd WEAR Dataset Challenge at HASCA 2026~\cite{wearchallenge,kaggle2026wear} turns the data into a window classification task. The 22 training participants (IDs 0 to 21) each performed 18 workout activities, and the test set comes from four new participants (IDs 22 to 25). The test recordings are cut into non-overlapping one-second windows, each window keeps only one randomly chosen limb, and the order of the windows is not given. Every window has 50 accelerometer samples and 15 frame embeddings of 768 values from a VideoMAEv2-Base model~\cite{wang2023videomaev2}, extracted by the organisers. The score is macro F1 over 19 classes (null plus 18 activities).

The challenge ran from 13 April to 5 July 2026, and the prize ranking used the private leaderboard at the technical report deadline~\cite{wearchallenge}. The work in this report was done afterwards, in September and October 2026, while the Kaggle leaderboard was still open. Our submissions are therefore not part of the prize ranking. The private scores of our submissions became visible when the Kaggle leaderboard closed in October 2026; we report them, but none of our choices used them. For the same reason, this report is not part of the challenge proceedings and does not follow their six-page format.

Without order, a window has to be classified from one second of data. With order, the same window is part of a 90-second bout, and its label can be read from its neighbours. The WEAR benchmark experiments~\cite{bock2024wear} use the full timeline, for example for temporal action localisation with ActionFormer~\cite{zhang2022actionformer} and TriDet~\cite{shi2023tridet}. The challenge removes the order, and our system recovers part of it from the data of each participant. It looks for the successor of each window, connects windows that look alike in video or IMU features, and labels all windows of a participant together under a soft constraint on the number of windows per activity. The system thus classifies windows in the context of a whole recording, and its results depend on how well this context is recovered.

The report has two aims. The first is to describe the system; Section~\ref{sec:method} and Appendix~\ref{app:impl} give the settings of every trained part, and the code is released (Section~\ref{sec:discussion}). The second is an error budget: which parts carry the score, and how far each part is from a ceiling. For this we use a participant-wise simulation of the test setting, a participant-level bootstrap for every comparison, seed replicates of one stacking chain, checks of the validation protocol and several oracles. Points raised in the competition forum, such as label errors in the training data and other ways to build the video graph, were checked in the same simulation and are reported where they apply.

This is a technical report on one challenge system, and two of its ideas were used in public notebooks of the challenge before this work: successor links and a per-activity count band (see the acknowledgements). Its contribution is empirical, in three parts:
\begin{enumerate}
\item a measurement of what the stacking stages and the count constraint add to a system for shuffled multimodal windows, including the cost of the constraint for activities with long bouts;
\item a comparison of seam distances for same-limb successors, which shows a gap between the quality of the links and the final score: the quadratic seam finds successors far more often than the alternatives we tried, but we could not detect an effect on macro F1;
\item an error budget with oracles, together with an analysis of the label problems in the training data and of the limits of the validation protocol.
\end{enumerate}
We do not compare the system with other challenge solutions under one protocol and make no claim that it is better than them; another team reports a higher public score of 0.928~\cite{mugoya2026post}. All results hold for the setting of the challenge, in which all windows of a participant are available at once and each activity is performed once; we make no claim about online or inductive recognition.

\section{Data, validation and metric}
\label{sec:data}

\subsection{Simulation of the test setting}
\label{sec:sim}

The training set has continuous recordings with labels per sample. We cut every recording into non-overlapping one-second tiles in the same way as the test set, which gives 69{,}326 tiles. A tile has 50 accelerometer samples per limb and the video frame embeddings $30t+8$ to $30t+22$ of second $t$, the frame range of the test windows. The tile label is the majority label of its 50 samples. Each tile keeps one limb, drawn at random with a fixed seed; if the drawn limb has missing samples in the tile, the first limb with complete data is used. Recordings stay whole, and each recording is decoded as one unit, like a test participant. The target scenario is the one of the challenge: all windows of a new participant are available together and are labelled offline, so the system is transductive. Participants 0 and 14 have two recordings each; for them, everything that the text describes per participant is done per recording.

The organisers released a metadata file for the test participants with the sessions of each participant, their durations and the number of activities per session: 9 and 8 plus a short third session with one activity for participants 22 and 23, and 9 and 9 for participants 24 and 25. Every test participant therefore performed each of the 18 activities once. For three of the four participants the number of test windows equals the summed session duration in seconds; participant 22 has 5{,}197 windows for 5{,}017 seconds of sessions. In the training set, five participants performed some or all activities twice. For participants 0, 2, 3 and 14 this is visible in the labels. Participant 10 has all 18 activities labelled before its last session, with 74 to 129 windows each, and that session is labelled null (Section~\ref{sec:labels}); there, the arg max of our stacked unary term gives 634 activity windows, 630 of them for the nine activities that most participants perform together in one session. We therefore count participant 10 among the participants with repeated activities. Every simulation result is reported for all 22 participants and for the 17 participants who performed each activity once, which is closer to the test setting. In the text, a number refers to all participants unless it is marked as single-performance.

\subsection{Folds and stacking protocol}
\label{sec:protocol}

All trainable parts of the system use the same partition of the 22 participants into five folds: \{1, 6, 11, 16, 21\}, \{0, 5, 10, 15, 20\}, \{4, 9, 14, 19\}, \{3, 8, 13, 18\} and \{2, 7, 12, 17\}. Each part is trained five times, each time without one fold, and predicts that fold. Later parts are trained on these out-of-fold outputs of earlier parts. Table~\ref{tab:protocol} lists what each part is trained on when the participants of fold $k$ are validated. The participant graphs of Section~\ref{sec:graph} are built from unlabelled features of one participant at a time, and features are standardised within a participant without labels, in the same way for each test participant. No label of a validation participant enters a graph, a feature or the decoder.

\begin{table}[t]
\caption{What each part of the system is trained or tuned on when the participants of fold $k$ are validated. OOF: out-of-fold.}
\label{tab:protocol}
\small
\setlength{\tabcolsep}{3pt}
\begin{tabular}{p{2.4cm}p{3.6cm}p{1.8cm}}
\toprule
Part & Trained or tuned on & Labels of fold $k$ \\
\midrule
Window models & windows of the other four folds & not used \\
Successor models & successor pairs of the other four folds & not used \\
Participant graphs & no training; unlabelled features of one participant & not used \\
Video PCA & no training; unlabelled frames of all training and test participants & not used \\
Stage 2 & OOF stage-1 outputs of the other four folds & indirectly$^{a}$ \\
Stage 3 & OOF stage-2 outputs of the other four folds & indirectly$^{a}$ \\
Fusion weights, decoder settings & pooled OOF outputs of all 22 participants$^{b}$ & in the selection \\
\bottomrule
\end{tabular}

\smallskip
\footnotesize
$^{a}$The OOF outputs of the other four folds come from models that were trained with fold $k$.\quad
$^{b}$The count bounds were also tuned partly on the public leaderboard.
\end{table}

The protocol is not nested. The out-of-fold stage-1 outputs of a training participant come from stage-1 models that were trained with the other folds, including fold $k$. Through these features, the labels of fold $k$ can influence the training data of the next stage. Section~\ref{sec:nested} measures the size of this effect with a reduced pipeline. The fusion weights, the mixing weight of the final unary term, the decoder settings and the choice of which variants to keep were made on the pooled out-of-fold predictions of all 22 participants, which adds a selection bias; Section~\ref{sec:nested} measures it for the decoder settings and the mixing weight. There is no automatic feature selection. The principal components of the video embeddings (Section~\ref{sec:method}), the fusion weights and the decoder settings were fitted or chosen once for all folds, not inside each fold. For the test set, each model is either refitted on all 22 participants or averaged over its five fold models (Appendix~\ref{app:impl}). The stacking stages are trained on out-of-fold outputs of models fitted on four folds, but at test time they receive outputs of models fitted on more data. Section~\ref{sec:nested} compares the confidence of the models on out-of-fold and test windows; the effect of this mismatch on the test score cannot be measured without test labels. The simulation scores in this report are therefore results of this particular protocol, and they are not fully independent estimates of the performance on new participants.

\subsection{Metric and uncertainty}

Macro F1 is computed on the pooled windows of all participants in a set: per-class F1 from one confusion matrix over the 19 classes, then the unweighted mean. This is the challenge metric applied to the simulation, and a participant with more windows has more weight in it. Appendix Table~\ref{tab:participants} gives the macro F1 of each participant separately, over the classes that occur in its labels or predictions. Each comparison between two variants gets a paired participant bootstrap. We resample the participants with replacement 2{,}000 times, recompute the pooled macro F1 of both variants on each resample, and report the 95\% percentile interval of the difference, together with the number of participants whose own macro F1 is higher with the variant. We resample participants because the windows of one participant are strongly dependent. When the interval of a difference contains zero, we call the difference not detectable; this does not show that the effect is zero.

\subsection{Training labels}
\label{sec:labels}

Problems in the training labels were reported in the competition forum. A public notebook by goodpjw2008~\cite{goodpjw2026notebook}, as summarised by habakan~\cite{habakan2026post}, compared the labels with the session table of the WEAR supplementary material. It found that the last 112\,s of participant 2 and the last session of participant 10 (1{,}419\,s) are labelled null although activities were performed, and that one session of participant 7 has a 10\,s delay. Our out-of-fold predictions agree on the first two. All 112 tiles of participant 2 are predicted as activity 2 (jogging with rotating arms), which is otherwise missing from that participant's labels, and the arg max of the stacked unary term marks 634 windows of the last session of participant 10 as activities (Section~\ref{sec:sim}). These 1{,}531 tiles are left out of every score in this report, so 67{,}795 tiles are scored, 46{,}015 of them from single-performance participants. We did not correct the delay of participant 7, because the summary we had did not say which session is affected; a 10\,s delay mislabels about ten windows at each label change of that session. Leaving participant 7 out of the score raises macro F1 from 0.9381 to 0.9405 (single-performance: from 0.9432 to 0.9470), and the reported scores include this participant. Section~\ref{sec:labelres} compares three ways of training with the two spans.

\section{Method}
\label{sec:method}

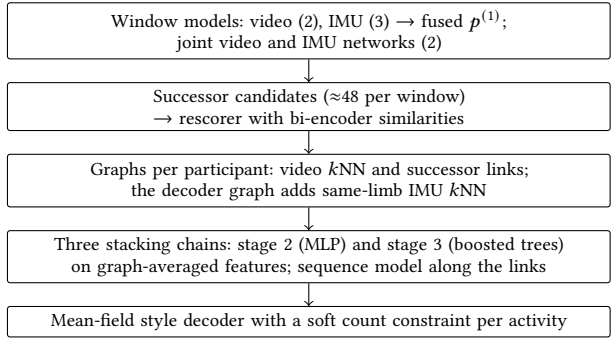
\begin{figure}[t]
\centering
\begin{tikzpicture}[font=\footnotesize, node distance=3mm,
  box/.style={draw, rounded corners=1pt, align=center, inner sep=2.5pt, text width=0.92\columnwidth}]
\node[box] (w) {Window models: video (2), IMU (3) $\rightarrow$ fused $p^{(1)}$;\\ joint video and IMU networks (2)};
\node[box, below=of w] (s) {Successor candidates ($\approx$48 per window) $\rightarrow$ rescorer with bi-encoder similarities};
\node[box, below=of s] (g) {Graphs per participant: video $k$NN and successor links;\\ the decoder graph adds same-limb IMU $k$NN};
\node[box, below=of g] (st) {Three stacking chains: stage 2 (MLP) and stage 3 (boosted trees)\\ on graph-averaged features; sequence model along the links};
\node[box, below=of st] (d) {Mean-field style decoder with a soft count constraint per activity};
\foreach \a/\b in {w/s, s/g, g/st, st/d} \draw[->] (\a) -- (\b);
\end{tikzpicture}
\caption{Overview of the system. Everything after the window models works on one participant at a time.}
\Description{Five boxes connected by arrows from top to bottom: window models, successor candidates and rescorer, participant graphs, three stacking chains, and the decoder with a count constraint.}
\label{fig:pipeline}
\end{figure}

Figure~\ref{fig:pipeline} shows the parts of the system. Appendix~\ref{app:impl} lists the settings of every trained part and states which parts are used in the final submission. The video transformer, the IMU CNN, the joint networks, the bi-encoders and the sequence model were trained on Kaggle GPUs with the full 768-dimensional frame embeddings. All other parts ran on a CPU and use a compressed copy of the embeddings, made once before all experiments: the first 224 principal components, which keep 93.1\% of the variance (mean cosine 0.977 between an embedding and its reconstruction), stored as 8-bit integers with one scale per component. The principal components were fitted once, without labels, on all frames of the 22 training participants and the four test participants. In simulation this fit includes the unlabelled frames of each validation participant, so it is transductive in the same way as the standardisation within a participant.

\subsection{Window models}
\label{sec:window}

Seven window classifiers give class probabilities from one window. Two use only video: an MLP on pooled frame features (mean, standard deviation, mean of the first three and of the last three frames, and mean absolute frame difference, standardised per participant) and a two-layer transformer over the 15 frames. Three use only the accelerometer: an MLP and a histogram gradient boosting model on 179 hand-made features per window, and a 1D CNN on the raw signal. The 179 features are 43 statistics for each of the three axes and the magnitude (mean, standard deviation, minimum, maximum and range, five quantiles, skewness and kurtosis, the standard deviations of the first and second differences, Hjorth mobility and complexity, zero-crossing rate, spectral entropy, centroid and peak frequency, log amplitudes of the first 15 Fourier bins, autocorrelations at six lags from 2 to 25 samples, the difference of the two half-window means and the mean absolute first difference), plus the three axis correlations, the unit vector of the mean acceleration and the mean squared magnitude. The IMU models are trained on all four limbs of every training tile, with the limb as an input. Two networks see both modalities; the second of them is trained on windows cut at random sub-second offsets. The fused stage-1 probability $p^{(1)}$ is a weighted mean of the log probabilities of the five single-modality models, with weight 0.2 for each video model, 0.2 for the IMU MLP, 0.3 for the IMU boosting model and 0.1 for the IMU CNN, chosen on out-of-fold predictions. The two joint networks enter the stacking stages (Section~\ref{sec:stacking}).

\subsection{Successor candidates and rescoring}
\label{sec:succ}

For every window $a$, which we call the anchor, we look for the window $b$ that followed it. A first-pass gradient boosting model scores about 48 candidates per anchor. Its features are the frame similarities that define the candidate lists with their ranks in the row of the anchor and the column of the candidate, the extrapolation distances of the accelerometer signal, the limb ids, the similarity of the stage-1 class probabilities, and the similarities of two linear end-to-start embeddings trained with an InfoNCE loss on the training folds (Appendix~\ref{app:impl}). The candidates are the union of three lists: the 30 windows whose first frames are most similar to the last frames of $a$, the 15 windows that best match ridge predictions of the next and the previous frames, and the 12 windows with the same limb as $a$ that best continue its accelerometer signal, judged by comparing the first sample of $b$ with a linear extrapolation of the last two samples of $a$. The true successor is in the candidate set for 99.2\% of same-limb pairs and 91.6\% of cross-limb pairs.

A rescorer then re-ranks the candidates of every anchor. It is a gradient boosting classifier~\cite{pedregosa2011sklearn} trained on all true successor pairs and a random 20\% of the other candidate pairs. Most of its features come from the $15\times15$ cosine matrix between the frames of $a$ and $b$, summarised along its diagonals and at the boundary, and from the decay of self-similarity inside each window. The others describe the position of the first-pass score inside its row and column: rank, gap to the best, margin to the second best and the score of the reverse pair. The later link models are a second rescoring round on top of these rescorer scores. The final one adds the limb ids of the two windows and the similarities of two bi-encoders, trained with an InfoNCE loss to match the end of a window to the start of its successor, one on the video frames and one on the accelerometer window of a single limb. Two further feature sets were tested and not kept: membership in a one-to-one Hungarian assignment~\cite{kuhn1955hungarian} of the score matrix, and the seam features of Section~\ref{sec:seam} (Table~\ref{tab:links}).

\subsection{Seam distance for same-limb pairs}
\label{sec:seam}

When $a$ and $b$ carry the same limb and $b$ follows $a$, the accelerometer signal continues across the boundary without a gap. The candidate search of Section~\ref{sec:succ} uses this with a linear extrapolation from $a$ only. For seam matching we use the last three samples of $a$ and the first three samples of $b$ and extrapolate both to the middle of the gap between them, half a sample after the end of $a$. With the backward differences $\nabla x$ and $\nabla^2 x$ of the last samples of $a$, and the forward differences $\Delta y$ and $\Delta^2 y$ of the first samples of $b$, Newton's interpolation formulas give
\begin{align}
\hat{x}_a &= x_{-1} + \tfrac{1}{2}\nabla x_{-1} + \tfrac{3}{8}\nabla^2 x_{-1},\\
\hat{y}_b &= y_{0} - \tfrac{1}{2}\Delta y_{0} + \tfrac{3}{8}\Delta^2 y_{0},
\end{align}
with $\nabla x_{-1} = x_{-1}-x_{-2}$, $\nabla^2 x_{-1} = x_{-1}-2x_{-2}+x_{-3}$, $\Delta y_0 = y_1-y_0$ and $\Delta^2 y_0 = y_2-2y_1+y_0$, where $x_{-1}$ is the last sample of $a$ and $y_0$ the first sample of $b$. The seam distance is $d(a,b)=\lVert \hat{x}_a-\hat{y}_b\rVert_2$ over the three axes. In the rescorer, the seam features are this distance, the same distance divided by the median absolute second difference of the two windows, its rank in the row of $a$ and in the column of $b$ among all same-limb windows, its gap to the row and column minimum, and three simpler variants (linear extrapolation from both sides, from $a$ only, and the difference of the end velocities). For cross-limb pairs these features are missing values.

\subsection{Participant graphs}
\label{sec:graph}

For each participant we build sparse graphs from three directed weight matrices. $W_v$ joins each window to its 10 nearest video neighbours, using the cosine similarity $s$ of the mean frame embedding after standardisation within the participant, with weight $e^{5(s-1)}$. $W_i$ does the same with the standardised IMU features, among windows with the same limb only. $W_\ell$ holds the successor candidates with weight $r^2$, where $r$ is the output of the link model. The decoder uses
\begin{equation}
A = (W_v + W_v^\top) + (W_i + W_i^\top) + 10\,(W_\ell + W_\ell^\top).
\end{equation}
A pair that appears in both directions gets the sum of both weights; this holds for 66\% of the video neighbour edges. Replacing the sum by the element-wise maximum changed macro F1 by $-0.0002$ (interval $-0.0006$ to $0.0004$). The mean $(W+W^\top)/2$ differs from the sum only by a constant factor, which the normalisation in the decoder removes. On average, the successor links carry 66\% of the total edge weight of a decoder graph, the IMU neighbours 19\% and the video neighbours 15\%. The candidate pairs are unique, so no pair is counted twice. The stacking stages use a second graph without IMU neighbours, $A_s = (W_v + W_v^\top) + \lambda\,(W_\ell + W_\ell^\top)$ with $\lambda = 10$ or 3 (Section~\ref{sec:stacking}).

\subsection{Stacking}
\label{sec:stacking}

The system has three stacking chains, each with a stage-2 MLP and a stage-3 gradient boosting model (Appendix~\ref{app:impl}). Both stages see one participant at a time through the row-normalised graph $H = D^{-1}A_s$. Stage 2 gets the window's own IMU features (standardised per participant and limb), its limb, 96 pooled video features (the means over the 15 frames of the first 64 principal components and the standard deviations of the first 32), the stage-1 log probabilities with their averages over one and two steps of $H$, and the log probabilities of the window models of its chain. For every limb separately, it also gets averages of the IMU features and of the mean IMU log probability over the neighbours that carry that limb, so a window sees the accelerometer signal of the other three limbs as it appears in its neighbours. Stage 3 gets the stage-1 and stage-2 log probabilities with their one- and two-step averages, the window-model log probabilities, the per-limb averages of the IMU log probabilities, the video features, the limb, and the labels of a decoder run on the stage-2 output, with their averages. This decoder is an earlier one of ours. On the mean of the stage-2 and stage-1 log probabilities it runs label propagation over the graph ($\alpha = 0.8$, 30 steps), clusters the windows into connected groups of about 25 with Ward linkage, and assigns groups to classes under count bounds of 70 and 125.

The three chains differ in their inputs. Chain~A uses the rescorer links (second row of Table~\ref{tab:links}) with $\lambda=10$, and its window models are the five single-modality models and the mean of the two joint networks. Chain~B adds a sequence model, a transformer trained on short walks along the best successor and predecessor candidates (eight steps each way). Chain~C uses the final link model with $\lambda=3$, includes the sequence model, and is trained with relabelled tiles (Section~\ref{sec:labelres}). Their stage-2 models reach out-of-fold window macro F1 of 0.877, 0.882 and 0.888, and their stage-3 models 0.918, 0.918 and 0.923. The final unary term is $\log p = 0.6\,\overline{\log p^{(3)}} + 0.4 \log p^{(1)}$, where $\overline{\log p^{(3)}}$ is the mean of the log probabilities of the three stage-3 models.

\subsection{Decoding with a soft count constraint}
\label{sec:decode}

All windows of a participant are labelled together. The update is in the style of mean-field inference~\cite{krahenbuhl2011densecrf} and close to label propagation~\cite{zhou2004consistency}, with the log of the neighbour average as the pairwise term:
\begin{equation}
q_i(c) \propto \exp\Big(\log p_i(c) + b_c + \beta \log\big(\tfrac{\sum_j A_{ij} q_j(c)}{\sum_j A_{ij}} + 10^{-3}\big)\Big).
\end{equation}
The decoder starts from $q_i = p_i$ and $b = 0$ and runs 30 iterations $t = 1,\dots,30$. After iteration $t$, for $6 \le t \le 29$, it counts $n_c = |\{i : \arg\max_k q_i(k) = c\}|$ and updates the bias of every activity class,
\begin{equation}
b_c \leftarrow b_c + 0.15\,[n_c < l] - 0.15\,[n_c > h],
\end{equation}
where $[\cdot]$ is 1 if the condition holds and 0 otherwise. The null class has no bias. The label of window $i$ is $\arg\max_c q_i(c)$ after iteration 30. This is a heuristic in the spirit of dual ascent on the constraints $l \le n_c \le h$, and it is not inference in a defined probabilistic model. The constraint is soft: the bias that enters the labels can reach at most $24 \times 0.15 = 3.6$ in absolute value, and the unary and pairwise terms can outweigh it. The constraint is used only at decoding. No model is trained with it as a loss term, and the stacking stages see the graph only through averaged features and the labels of the earlier decoder (Section~\ref{sec:stacking}). In the final simulation no class bias reversed its direction more than once, and 73.5\% of the predicted (participant, activity) counts of single-performance participants ended inside $[l,h]$. The final submission uses $l=75$, $h=95$ and $\beta=1.5$. These bounds are decoding parameters, and Section~\ref{sec:counts} compares them with the true counts. In simulation, the five participants who performed activities twice get $h=125$ and $l=\min(l,80)$; the test metadata rule this case out for the test participants.

\section{Results}

\subsection{Build-up of the system}
\label{sec:buildup}

Table~\ref{tab:stages} adds the parts one at a time. Stacking is the largest step. The stacked unary term alone reaches 0.9236, against 0.7724 for the fused window models, and the final system is 0.0505 above stage-1 fusion with the same decoder (interval 0.0419 to 0.0594, better for all 22 participants). The decoder adds 0.0144 on top of the stacked unary term (0.0103 to 0.0191). The count constraint is the largest part of that: without it the decoder reaches 0.9269, and with it 0.9381, a gain of 0.0112 (0.0061 to 0.0167) that holds for 18 of 22 participants (single-performance: 0.0122, 0.0058 to 0.0196, 15 of 17). Removing the successor links from the decoder costs 0.0067 (0.0050 to 0.0087, all 22 participants). Removing the video or the IMU neighbours costs 0.0005 and 0.0007, with intervals that contain zero. The stacking stages also use the video neighbours and the successor links, so for these two edge types the decoder ablation does not show their whole contribution; the IMU neighbour edges enter only the decoder.

\begin{table}[t]
\caption{Build-up, ablations and oracles in simulation (macro F1 for all participants and for single-performance participants). The indented rows are variants of the final system. Oracles use information that is not available at test time.}
\label{tab:stages}
\small
\setlength{\tabcolsep}{4pt}
\begin{tabular}{lcc}
\toprule
& All & Single \\
\midrule
Video transformer (best video-only model) & 0.5777 & 0.6035 \\
IMU gradient boosting (best IMU-only model) & 0.6139 & 0.6208 \\
Mean of the two joint networks & 0.6790 & 0.6956 \\
Fused window models $p^{(1)}$, arg max & 0.7724 & 0.7816 \\
$p^{(1)}$ + decoder & 0.8876 & 0.8923 \\
Stacked unary term, arg max & 0.9236 & 0.9279 \\
Stacked + decoder, no count constraint & 0.9269 & 0.9309 \\
Stacked + decoder (final) & 0.9381 & 0.9432 \\
\quad conservative estimate (Section~\ref{sec:nested}) & 0.9355 & 0.9403 \\
\quad without successor links & 0.9313 & 0.9367 \\
\quad without video neighbours & 0.9376 & 0.9429 \\
\quad without IMU neighbours & 0.9374 & 0.9425 \\
\midrule
Oracle: activity group (session) known & 0.9407 & 0.9454 \\
Oracle: true successor links & 0.9446 & 0.9493 \\
Oracle: true order, Viterbi smoothing & 0.9495 & 0.9504 \\
Oracle: true segments, mean stacked unary & 0.9727 & 0.9714 \\
Oracle: true segments, mean stage-1 unary & 0.9385 & 0.9421 \\
\bottomrule
\end{tabular}
\end{table}

The participant bootstrap does not cover the variation between training runs. To estimate it, we retrained the stage-2 model of chain~C with two other seeds and kept everything else fixed; stage 3 has no random parts. The stage-3 window macro F1 of chain~C was 0.9228, 0.9204 and 0.9214, and the final system scored 0.9381, 0.9374 and 0.9380 (single-performance: 0.9432, 0.9428 and 0.9430). The submitted version is the best of the three, by up to 0.0007, which is as large as several of the effects reported below. On the test set, the two retrained versions changed 68 and 79 of the 12{,}234 predicted labels.

\subsection{Successor links and the seam distance}
\label{sec:links}

Table~\ref{tab:seam} compares seam distances on all 17{,}601 same-limb successor pairs of the training simulation. Each anchor is ranked against all other windows with the same limb in its recording. Extrapolating from both sides is better than from one side, and the quadratic form is the best of the forms we tried. It ranks the true successor first for 83.9\% of anchors (interval 82.8 to 84.9\%), and it beats one-sided linear extrapolation for each of the 22 participants. Dividing the distance by the typical second difference of the two windows lowers the top-1 rate to 77.3\%, and a cubic term lowers it to 74.8\%; we assume that higher differences amplify sensor noise. The difference of the end velocities alone says almost nothing about the order (1.1\%), and an AR(6) model fitted to each window and run forward and backward for four samples reaches 45.7\%, close to the rate of about 50\% that Mugoya~\cite{mugoya2026post} reports for an AR(6) seam score. The quadratic seam is best for the stretches (87 to 95\%) and sit-ups (93\%) and worst for the jogging variants (64 to 76\%) and burpees (73\%), the most dynamic activities. Across a label change it is almost as good as inside a segment (82.0\% for the 384 pairs whose two windows carry different labels).

\begin{table}[t]
\caption{Same-limb successor ranking. Each anchor is ranked against all other windows with the same limb in its recording (17{,}601 anchors, median 719 candidates). Interval: 95\% participant bootstrap of the top-1 rate. Participants: range of the per-participant top-1 rate.}
\label{tab:seam}
\footnotesize
\setlength{\tabcolsep}{3pt}
\begin{tabular}{lccc}
\toprule
Seam distance & Top-1 (interval) & Top-5 & Participants \\
\midrule
Last sample of $a$ vs.\ first of $b$ & 0.595 (0.577--0.612) & 0.818 & 0.499--0.666 \\
End-velocity difference & 0.011 (0.009--0.013) & 0.048 & 0.005--0.021 \\
Linear, from $a$ only & 0.685 (0.666--0.703) & 0.878 & 0.569--0.744 \\
Linear, both sides & 0.790 (0.775--0.803) & 0.922 & 0.705--0.836 \\
Quadratic, both sides (ours) & 0.839 (0.828--0.849) & 0.938 & 0.781--0.882 \\
Quadratic, normalised & 0.773 (0.757--0.787) & 0.917 & 0.670--0.828 \\
Cubic, both sides & 0.748 (0.732--0.763) & 0.897 & 0.665--0.803 \\
AR(6), both sides & 0.457 (0.440--0.472) & 0.681 & 0.375--0.519 \\
\bottomrule
\end{tabular}
\end{table}

Table~\ref{tab:links} lists the link models in the order we built them. The F1 columns show the full system when only the link scores in the decoder are changed; the stacking stages are those of the final system. Top-1 accuracy rose from 0.534 to 0.667. Macro F1 rose by 0.0023 from the first-pass links to the final link model (interval 0.0013 to 0.0035). The seam features moved the top-1 rate of same-limb pairs from 0.852 to 0.920 and changed macro F1 by $-0.0001$ ($-0.0006$ to $0.0005$). Rebuilding stage 2 and the stage-3 model of chain~C with the seam links gave 0.9383, a difference of 0.0002 ($-0.0007$ to $0.0011$). Mugoya~\cite{mugoya2026post} also found a small effect of seam features in another pipeline. Pruning the links to the best one per anchor, the best three, or pairs with $r>0.1$ changed macro F1 by $-0.0006$ to $0.0000$, with intervals that contain zero, and keeping only mutually best pairs cost 0.0018 (0.0009 to 0.0029).

\begin{table}[t]
\caption{Link models. Each indented row adds its features to the nearest row above it with less indentation, so the Hungarian variant and the video bi-encoder both extend the rescorer. Top-1: share of anchors whose highest-scored candidate is the true successor, for all, same-limb and cross-limb pairs. F1: macro F1 of the final system with only the decoder's link scores changed.}
\label{tab:links}
\small
\setlength{\tabcolsep}{3.5pt}
\begin{tabular}{lccccc}
\toprule
& \multicolumn{3}{c}{Top-1} & \multicolumn{2}{c}{Macro F1} \\
Links & all & same & cross & all & single \\
\midrule
First-pass pair model & 0.534 & 0.795 & 0.446 & 0.9357 & 0.9406 \\
+ rescorer & 0.568 & 0.812 & 0.485 & 0.9364 & 0.9416 \\
\quad + Hungarian & 0.586 & 0.810 & 0.510 & 0.9361 & 0.9412 \\
\quad + video bi-encoder & 0.628 & 0.809 & 0.567 & 0.9376 & 0.9425 \\
\quad\quad + IMU bi-encoder (final) & 0.641 & 0.852 & 0.569 & 0.9381 & 0.9432 \\
\quad\quad\quad + seam & 0.667 & 0.920 & 0.581 & 0.9380 & 0.9434 \\
\midrule
True successors & 1 & 1 & 1 & 0.9446 & 0.9493 \\
\bottomrule
\end{tabular}
\end{table}

We see two reasons for the small effect of better links. About three quarters of the successor pairs are cross-limb (one random limb out of four), and cross-limb top-1 is 0.57 for the final link model. And once a window is linked to a few windows of the same bout, more correct links add little to what the stacked unary term and the other edges already carry. Even the true links add only 0.0065.

\subsection{The count constraint}
\label{sec:counts}

Appendix Table~\ref{tab:classes} gives, for every activity, the distribution of its number of windows over the 17 single-performance training participants. The class medians range from 87 (bench-dips) to 102 (sit-ups, complex), the 10th percentiles from 75 to 94 and the 90th percentiles from 95 to 114. The bounds $l=75$ and $h=95$ do not describe the range of these counts: 44\% of the true (participant, activity) counts are above 95, 21\% above 100 and 4\% below 75, with a range of 51 to 144. For the single-performance participants, the constraint lowered the mean absolute difference between the predicted and the true number of windows of an activity from 8.4 to 5.4, and Table~\ref{tab:classes} gives this difference per class. For the five classes whose median is above 95, the constraint makes the predicted counts lower than the true ones by 4.4 windows on average, against 0.1 without it. Bounds taken per class from the 10th and 90th percentiles of the training folds scored 0.9375, and from the 25th and 75th percentiles 0.9380. Neither differs from one band for all classes ($-0.0005$, interval $-0.0022$ to $0.0012$, and $-0.0001$, $-0.0016$ to $0.0016$).

The count constraint changed the label of 1{,}904 windows: 1{,}226 became correct, 607 became wrong and 71 changed from one wrong label to another. The largest net gains were for null ($+368$), push-ups complex ($+143$) and sit-ups ($+114$); the largest net loss was for sit-ups complex ($-94$), whose median of 102 windows lies above $h$.

Table~\ref{tab:sens} shows the sensitivity to $l$ and $h$ at $\beta=1.5$. The scores form a plateau for $l$ between 75 and 85 and $h$ between 95 and 100 (0.9380 to 0.9382 for all participants, 0.9430 to 0.9433 for single-performance ones). At $h=90$ they drop by 0.0019 to 0.0026 (single-performance: 0.0027 to 0.0037), and above $h=100$ they decline slowly. The coupling weight matters about as much: at $l=75$ and $h=95$, $\beta=1$ and $\beta=2$ lost 0.0023 and 0.0019 (single-performance: 0.0013 and 0.0033). The step size and the number of iterations act mainly through the largest bias a class can reach, the step times the number of iterations after the fifth, which is 3.75 in the final setting. Settings with a largest bias of 3.0 or 3.5 scored within 0.0008 of it, 2.25 and 2.5 lost 0.0013 and 0.0014, 1.5 lost 0.0031, 5.0 and 5.25 lost 0.0031 and 0.0051, and 7.0 lost 0.0098. A large bias presumably overrides the evidence for bouts that are longer than $h$.

\begin{table}[t]
\caption{Sensitivity of simulated macro F1 to the count bounds $l$ and $h$ ($\beta=1.5$, final unary term), for all participants and for single-performance participants. Final setting in bold.}
\label{tab:sens}
\small
\setlength{\tabcolsep}{2.6pt}
\begin{tabular}{llcccccc}
\toprule
& & $h=90$ & 95 & 100 & 105 & 110 & 125 \\
\midrule
All & $l=70$ & 0.9348 & 0.9374 & 0.9375 & 0.9368 & 0.9358 & 0.9346 \\
& $l=75$ & 0.9356 & \textbf{0.9381} & 0.9380 & 0.9373 & 0.9364 & 0.9353 \\
& $l=80$ & 0.9357 & 0.9380 & 0.9381 & 0.9374 & 0.9364 & 0.9355 \\
& $l=85$ & 0.9362 & 0.9381 & 0.9382 & 0.9377 & 0.9367 & 0.9360 \\
\midrule
Single & $l=70$ & 0.9391 & 0.9428 & 0.9429 & 0.9419 & 0.9404 & 0.9388 \\
& $l=75$ & 0.9396 & \textbf{0.9432} & 0.9430 & 0.9421 & 0.9407 & 0.9392 \\
& $l=80$ & 0.9398 & 0.9431 & 0.9432 & 0.9422 & 0.9407 & 0.9395 \\
& $l=85$ & 0.9405 & 0.9432 & 0.9433 & 0.9427 & 0.9411 & 0.9402 \\
\bottomrule
\end{tabular}
\end{table}

The choice between $h=95$ and $h=100$ cannot be made from the simulation (difference 0.0001, interval $-0.0006$ to $0.0009$; single-performance 0.0002, $-0.0008$ to $0.0012$). It was made on the public leaderboard, where $h=95$ scored 0.0011 higher (Table~\ref{tab:subs}). The best cell of Table~\ref{tab:sens}, $l=85$ and $h=100$, scores 0.9382, so a choice made on the simulation alone would give simulation scores within 0.0001 of those reported here; the number that depends on the leaderboard choice is the public score of the final submission. The public leaderboard is computed on part of the test set and was used several times for such choices, so the final public score is not an independent estimate. Simulation (all participants) and public leaderboard moved in the same direction in five of the eight steps of Table~\ref{tab:subs}; they disagree for the new stage 3 with $l=80$, for the seam links and for $l=75$. The private scores in Table~\ref{tab:subs} became visible only after all choices had been made. Before the public leaderboard guided the choice of bounds (rows 1 and 2), the private scores were 0.0007 and 0.0008 above the public ones; from row 3 on, they were 0.0007 to 0.0026 below. Over the nine rows, the private score rose by 0.0008, against 0.0039 on the public leaderboard and 0.0049 in simulation, and the direction of its changes agreed with the simulation in four of the eight steps and with the public leaderboard in five. Most of the leaderboard gain of the tuned bounds therefore did not carry over to the private part of the test set. The selected submission (row 9) scored 0.92308 on the private leaderboard and placed 18th when the leaderboard closed. After the final choice, we submitted one more configuration (row 10): chain~C with the mislabelled spans left out, decoded with $l=85$ and $h=100$, the best cell of Table~\ref{tab:sens}. It scored 0.9375 in simulation (single-performance 0.9429; difference to the final system $-0.0005$, interval $-0.0022$ to $0.0011$), 0.92528 on the public leaderboard, 0.00014 below the final submission, and 0.92473 on the private leaderboard, 0.00165 above it. A configuration that uses neither the leaderboard choice of the bounds nor the pseudo-labels thus did as well on the public part and slightly better on the private part; with four test participants, we cannot tell whether this difference is more than noise. The first two steps of Table~\ref{tab:subs} changed 97 and 225 of the 12{,}234 test labels, and each later step changed 25 to 77, about as many as a new seed for one stage-2 model (Section~\ref{sec:buildup}). In the final submission, the four test participants have 88 to 95 predicted windows per activity on average (range 75 to 126).

\begin{table}[t]
\caption{Kaggle submissions leading to the final system (rows 1 to 9), in the order they were made, and one later check (row 10, chain C with the mislabelled spans left out, made after the final choice). Row 1 uses the stage-3 models of chains A and B and of a chain with the video bi-encoder links; row 2 replaces the last by chain C trained with the original labels, and row 9 by chain C with pseudo-labels. Each row is evaluated in simulation with its own unary term. A step is compared with the configuration it changes: rows 2 and 3 with the row above, rows 4 to 6 with row 3, rows 7 and 8 with row 6, and row 9 with row 8. Sim: simulation macro F1 for all and for single-performance participants. Public and Private: public and private leaderboard; the private scores were seen only after the leaderboard closed.}
\label{tab:subs}
\small
\setlength{\tabcolsep}{2pt}
\begin{tabular}{rlccccc}
\toprule
& Change & $l$/$h$/$\beta$ & \multicolumn{2}{c}{Sim} & Public & Private \\
& & & all & single & & \\
\midrule
1 & Base & 70/125/2 & 0.9332 & 0.9371 & 0.92154 & 0.92225 \\
2 & New stage 3, $l=80$ & 80/125/2 & 0.9348 & 0.9384 & 0.92129 & 0.92211 \\
3 & Lower $h$ and $\beta$ & 80/100/1.5 & 0.9367 & 0.9430 & 0.92317 & 0.92063 \\
4 & Seam links & 80/100/1.5 & 0.9373 & 0.9433 & 0.92261 & 0.92191 \\
5 & $h=105$ & 80/105/1.5 & 0.9363 & 0.9424 & 0.92207 & 0.91948 \\
6 & $h=95$ & 80/95/1.5 & 0.9373 & 0.9438 & 0.92424 & 0.92161 \\
7 & $h=92$ & 80/92/1.5 & 0.9360 & 0.9420 & 0.92190 & 0.92037 \\
8 & $l=75$ & 75/95/1.5 & 0.9368 & 0.9433 & 0.92501 & 0.92308 \\
9 & Pseudo-labels, stage 3 & 75/95/1.5 & 0.9381 & 0.9432 & 0.92542 & 0.92308 \\
\midrule
10 & Spans left out & 85/100/1.5 & 0.9375 & 0.9429 & 0.92528 & 0.92473 \\
\bottomrule
\end{tabular}
\end{table}

The effect of the count constraint was similar for three limb draws. With stage-1 fusion and a decoder without successor links, the three draws gave window-level macro F1 of 0.7705 to 0.7724, and decoded scores of 0.8351 to 0.8387 with the constraint against 0.8015 to 0.8059 without it.

\subsection{Training with the mislabelled spans}
\label{sec:labelres}

The two mislabelled spans can be used for training in three ways: with their original labels, left out of training, or relabelled. For relabelling, the 112 tiles of participant 2 were set to activity 2, which the session table supports. The tiles of participant 10 were set to the out-of-fold output of our decoder in the configuration of row 3 of Table~\ref{tab:subs}, so they are pseudo-labels. That decoder treated participant 10 as a single-performance participant, and its upper bound pushed windows of the repeated session towards null: it marked 415 of the 1{,}419 tiles as activities, against 634 for the arg max of the stacked unary term, so the pseudo-labels probably under-count the activities of that session. We retrained stage 2 and the stage-3 model of chain~C in each of the three ways and kept everything else fixed. The final system scored 0.9368 with the original labels, 0.9374 with the spans left out and 0.9381 with the pseudo-labels (single-performance: 0.9433, 0.9428 and 0.9432). All three pairwise differences have bootstrap intervals that contain zero; pseudo-labels against original labels, for example, gave $+0.0013$ ($-0.0005$ to $0.0035$). The final submission used the pseudo-labelled model, and we do not claim that this choice improved the result. As long as the pseudo-labels have no independent check, the variant with the spans left out (0.9374; single-performance 0.9428) is the conservative estimate for this system. Because the pseudo-labels come from models that saw the other participants, the simulation score of this variant may also be slightly optimistic. The comparison retrains only stage 2 and the stage-3 model of chain~C. The window models, the link models and chains A and B were trained with the original labels in all three cases, so it measures the effect of the label policy on one stacking chain and not end to end. The pseudo-labels of participant 10 also carry the bias of the count constraint towards null into the training data of chain~C.

\subsection{Error budget}
\label{sec:budget}

The oracles of Table~\ref{tab:stages} bound what better context could give with our unary term; they are not bounds for the task itself. Knowing which of the two groups of nine activities a window belongs to, which for most participants means knowing its session, adds 0.0026. The true successor links inside our decoder add 0.0065. Viterbi smoothing along the true order, with one switch penalty (the better of 6 and 10 nats on this data), adds 0.0114. The true order gives both the neighbours of a window and the segment boundaries, so these gains cannot be assigned to a single part of the system. The strongest oracle knows the true segments of each recording, including the null segments, and gives every window of a segment the arg max of the mean unary log probability of the segment. It reaches 0.9727 with the final unary term. With the stage-1 unary term, which has no graph context, it reaches 0.9385, about the score of the full system. The stacked unary term already contains graph context, so this oracle does not separate naming a bout from grouping its windows cleanly. It does show that, with the true segments, the error of the final unary term in macro F1 would fall from 0.062 to 0.027.

We tried to recover the session of each window without labels. Propagating session evidence from confident activity windows over the graph gave the right session for 93\% of the windows, but in an earlier configuration a decoder that used this estimate scored the same as one without it (0.9302 against 0.9303).

The final system makes 4{,}192 errors in simulation (6.2\% of the scored tiles). Of these, 2{,}166 are activity windows labelled as null, 1{,}143 are null windows labelled as an activity and 883 are confusions between two activities, so 79\% involve the null class. By position, 1{,}566 errors (37\%) lie within two windows of a segment boundary and 2{,}142 (51\%) farther inside an activity segment; 1{,}691 errors (40\%) are at least ten windows away from a boundary. Another 260 lie inside a null segment between two different activities, and 224 inside a null segment between two parts of the same activity, which we call a pause. Pauses cover 10{,}859 windows, and 95.5\% of them are predicted as null. Appendix Table~\ref{tab:classes} lists precision, recall and F1 per class. Recall is lowest for sit-ups (0.878), lunges (0.899) and push-ups complex (0.902). The missed windows of sit-ups, bench-dips and the stretches are mostly predicted as null (73\% for sit-ups, 97 to 100\% for bench-dips and the shoulder and hamstring stretches), while push-ups and push-ups complex are mostly confused with each other. One explanation for the losses to null is that participants slow down or rest within these activities without a labelled break. This is a hypothesis; we did not check it against the raw videos.

Per-participant macro F1 ranges from 0.796 (participant 4) to 0.974 (participants 17 and 21; Appendix Table~\ref{tab:participants}). The mean of the per-participant scores is 0.938 with a standard deviation of 0.045 (single-performance: 0.943 and 0.045), so one pooled score describes the participants only roughly. The first eleven participants score 0.919 together and the last eleven 0.958. Our simulation (0.938) is above the public (0.925) and the private (0.923) leaderboard scores, gaps of 0.013 and 0.015 that are small against the spread between participants.

\subsection{Variants that did not help}
\label{sec:variants}

Several variants changed the simulation score by less than the bootstrap can resolve, or lowered it. Lowering the null log probability by 0.25 before decoding gave $+0.0004$ ($-0.0003$ to $0.0012$), and offsets up to 1.0 gave similar values. Removing the 10, 30 or 50 leading principal components of the video embeddings before the neighbour search, as proposed by habakan~\cite{habakan2026post} for a different graph, changed the score by at most $-0.0003$. More weight on IMU-only probabilities in the unary term, which Mugoya~\cite{mugoya2026post} found helpful in another pipeline, lowered it by 0.0009 (0.0004 to 0.0015) at weight 0.1. A kNN prior from the 20 nearest labelled training windows in video space, a one-step form of label propagation from the training set, lowered the score by 0.0004 at weight 0.1 (0.0000 to 0.0008) and by 0.0014 at weight 0.25 (0.0008 to 0.0021); on its own it reaches 0.424. In earlier configurations, a per-participant linear discriminant analysis trained on confident decoder outputs did not change the decoded score, and Viterbi decoding along chains built from a one-to-one assignment of the link scores reached only 0.918. Bounds that adapt to the data did not help either. Bounds per class, set inside the folds to the 10th and 90th percentiles of the true counts of the single-performance training participants, gave $-0.0005$ ($-0.0022$ to $0.0012$), and the 25th and 75th percentiles gave $-0.0001$ ($-0.0016$ to $0.0016$). Bounds scaled for each participant by $m$, the number of windows whose unary arg max is not null divided by 18, gave $-0.0020$ ($-0.0044$ to $-0.0003$) for $[0.8m, m]$ and $-0.0008$ and $-0.0010$ for $[0.85m, 1.05m]$ and $[0.9m, 1.1m]$. Averaging the log probabilities of the three seed replicates of chain~C (Section~\ref{sec:buildup}) before decoding gave $-0.0003$ ($-0.0010$ to $0.0004$), so we did not combine submissions by voting either.

\subsection{Checks of the validation protocol}
\label{sec:nested}

To measure the effect of the non-nested protocol of Section~\ref{sec:protocol}, we built a reduced two-stage pipeline that can be retrained quickly. Stage 1 is a gradient boosting model (50 iterations, trained on a random half of the training windows) on the IMU features of the drawn limb, the limb and 32 principal components of the video embeddings. Stage 2 is a gradient boosting model on the stage-1 log probabilities and their one- and two-step averages over the video neighbour graph. In the plain version, stage 2 is trained on ordinary out-of-fold stage-1 outputs, as in our system. In the nested version, the stage-1 outputs of the training participants come, for every outer fold, from an inner cross-validation over the remaining four folds, so no label of the outer validation fold reaches the training data of stage 2. Stage 1 alone scored 0.628. Stage 2 scored 0.7395 with the plain protocol and 0.7412 with the nested one (difference $-0.0017$, interval $-0.0098$ to $0.0067$). After decoding with the count constraint and without links, the plain version scored 0.8052 and the nested version 0.8157 (difference $-0.0105$, interval $-0.0195$ to $-0.0020$). In this reduced pipeline the plain protocol did not overstate the result. The nested version has a confound of its own: its stage 2 is trained on outputs of stage-1 models fitted on three folds and applied to outputs of models fitted on four, and we did not separate this effect from the nesting. The check does not cover the link models or the third stage of the full system.

The decoder settings and the mixing weight of the final unary term were chosen on the pooled out-of-fold predictions. To measure the optimism of this choice, we repeated it inside the folds. For every outer fold, the setting with the best pooled macro F1 on the other four folds was taken from a grid of mixing weights (0.5, 0.6 and 0.7), lower bounds (70 to 85), upper bounds (90 to 110) and coupling weights (1.25 to 2), and it was applied to the held-out fold. This gave 0.9359 against 0.9381 (difference $-0.0022$, interval $-0.0051$ to $-0.0003$; single-performance 0.9403 against 0.9432). The folds chose different settings, with lower bounds of 75 to 85, upper bounds of 95 or 100, mixing weights of 0.6 or 0.7 and coupling weights of 1.5 or 1.75. With the mislabelled spans also left out of training, the score is 0.9355 (single-performance 0.9403). We regard this as the least optimistic estimate of the system that we have. It uses no leaderboard information, but it still contains the optimism of the non-nested stacking and of the design choices made earlier on the same data.

The principal components of the video embeddings were fitted on the frames of all participants, including the validation and the test participants. Refitting them would mean retraining every part that uses the compressed embeddings, which we did not do. As a check of the decoder's video graph, we computed a second projection to 128 components from the compressed embeddings and decoded with it. Fitted on the training folds only, separately for each fold, it scored 0.9376; fitted on all 22 training participants, so with the unlabelled frames of the validation fold, 0.9379; and fitted on the training and the test participants, also 0.9379 (difference to the first, $-0.0004$, interval $-0.0007$ to $-0.0001$). The unlabelled frames of the validation participants thus help the decoder graph by a small amount, and the test frames add nothing measurable. The effect on the link models and the stacking stages, which also use the compressed embeddings, was not measured.

At test time the stacking stages receive outputs of window models that were refitted on all participants or averaged over the fold models. On the test windows, the window models were less confident than on the out-of-fold windows: the mean largest class probability was 0.596 against 0.636 for the stage-1 fusion, 0.524 against 0.635 for the video transformer and 0.710 against 0.720 for the IMU boosting model. The three stage-3 models were equally confident on both (0.974 to 0.976 against 0.975 to 0.977). The drop is spread over the classes: for the stage-1 fusion, windows with null as arg max had a mean largest probability of 0.511 on the test windows against 0.545 out of fold, and windows with an activity 0.664 against 0.691. On the out-of-fold windows, the expected calibration error (15 bins) is 0.154 for the stage-1 fusion and 0.031 for the final unary term; on the test windows it cannot be computed. The test participants differ from the training participants, so this check cannot separate the effect of refitting from that of the participants. A test-time construction that matches the training one, with fold-averaged outputs for every window model, was not run, and its effect on the test score could not be measured without test labels. Table~\ref{tab:checks} collects the scores of the final system under the variants of this section and of Sections~\ref{sec:labels} and~\ref{sec:labelres}.

\begin{table}[t]
\caption{Simulation macro F1 of the final system under variants of the validation protocol, for all participants and for single-performance participants.}
\label{tab:checks}
\small
\setlength{\tabcolsep}{3pt}
\begin{tabular}{p{4.75cm}cc}
\toprule
Variant & All & Single \\
\midrule
Final system, main protocol & 0.9381 & 0.9432 \\
Decoder settings and mixing weight chosen inside the folds & 0.9359 & 0.9403 \\
\quad and mislabelled spans left out of training & 0.9355 & 0.9403 \\
Mislabelled spans left out of training & 0.9374 & 0.9428 \\
Stage 2 of chain C with another seed & 0.9374 & 0.9428 \\
\quad with a third seed & 0.9380 & 0.9430 \\
Decoder video graph, projection fitted on training folds & 0.9376 & 0.9429 \\
\quad fitted on all training participants & 0.9379 & 0.9432 \\
\quad fitted on training and test participants & 0.9379 & 0.9432 \\
Participant 7 left out of the score & 0.9405 & 0.9470 \\
Mean (SD) of per-participant macro F1 & 0.938 (0.045) & 0.943 (0.045) \\
\bottomrule
\end{tabular}
\end{table}

\section{Discussion}
\label{sec:discussion}

Most of the remaining error of our system comes from grouping windows into bouts. With the true segments, the final unary term would reach 0.973. Half of the errors of the system lie inside activity segments, and two thirds of these are activity windows predicted as null. The count constraint helps because it adds a statement about whole bouts that local edges cannot make: each activity appears once and lasts about 90 seconds. This is a property of the WEAR protocol and of the challenge, and in recordings where activities repeat or vary much in length the bounds would have to be set differently or dropped. The system is also offline and transductive: it needs all windows of a participant before it labels any of them. Its bounds are decoding parameters that lie below many true counts, and they cost windows for the classes with long bouts. Better local links stopped helping after the IMU bi-encoder. A model of whole bouts, or of the pauses inside a bout, seems a more promising direction than another link model.

Three accelerometer samples at each end of a window are enough to find the next same-limb window among about 720 candidates in 84\% of cases. If future editions want to test window classification without context, gaps between the released windows would remove this shortcut, as habakan also suggests~\cite{habakan2026post}.

All ablations of the full system use one limb draw and one fold partition. The participant bootstrap covers the variation between participants. The variation between training runs was measured only for stage 2 of one chain; the GPU models of stage 1 and the link models were trained once, and repeated runs of the whole system, other limb draws for it and other fold partitions were beyond our compute. Differences of a few thousandths should be read with this in mind. The participants differ far more than the variants do: per-participant macro F1 ranges from 0.796 to 0.974, so a score on four test participants depends strongly on who they are. The count bounds were tuned partly on the public leaderboard, and most of that gain did not carry over to the private leaderboard, where a submission with bounds taken from the simulation scored 0.0017 higher. The relabelling of participant 10 uses our own predictions. Choosing the decoder settings inside the folds cost 0.002, and the other choices made on the pooled out-of-fold predictions were not checked in this way. The nested check covers a reduced pipeline only, the principal components of the video embeddings were fitted on all participants without labels, and the stacking stages see outputs of differently trained models in simulation and on the test set.

The code is available at \url{https://github.com/soheilkooklan/wear-seam-decoder} (release v1.1). It has two parts. The first is a clean implementation of the seam distance, the participant graph and the decoder in Python with NumPy and SciPy, with tests and with the unary scores and graphs of two training participants. On these two participants (5{,}747 windows) it gives exactly the labels of the final system; this checks the decoder, not the whole pipeline. The second holds the research scripts that produced the results, as they were used, for the window models, the link models, the stacking stages and the analyses of this report. They come with the order of the steps and the library versions, but they need the competition data and leave out a few short interactive steps, so they document the pipeline without forming a one-command reproduction; the trained models are not released.

\begin{acks}
We thank Marius Bock and the WEAR organisers for the dataset and the challenge. The label errors in the training data were found by goodpjw2008~\cite{goodpjw2026notebook}. Successor links and a per-activity count band appear in public notebooks of the challenge by Hanh Tran and Koushik Rudra, as credited in~\cite{mugoya2026post} and~\cite{habakan2026post}; our first count band used a similar range. AI assistants were used in this work: Claude (Anthropic) debug code, edit the text, and ChatGPT (OpenAI) and Grok (xAI) to review the paper and the code. The author directed the work, ran the GPU training on Kaggle, chose the submitted configurations, checked the numbers against the logs and takes full responsibility for the content.
\end{acks}

\balance
\bibliographystyle{ACM-Reference-Format}
\bibliography{refs}


\begin{thebibliography}{14}

%%% except for \shownote{} and \showURL{}.  The latter does
%%% not use final punctuation, in order to avoid confusing it with
%%% the Web address.

\ifx \showCODEN    \undefined \def \showCODEN     #1{\unskip}     \fi
\ifx \showISBNx    \undefined \def \showISBNx     #1{\unskip}     \fi
\ifx \showISBNxiii \undefined \def \showISBNxiii  #1{\unskip}     \fi
\ifx \showISSN     \undefined \def \showISSN      #1{\unskip}     \fi
\ifx \showLCCN     \undefined \def \showLCCN      #1{\unskip}     \fi
\ifx \shownote     \undefined \def \shownote      #1{#1}          \fi
\ifx \showarticletitle \undefined \def \showarticletitle #1{#1}   \fi
\ifx \showURL      \undefined \def \showURL       {\relax}        \fi
% invisible to TeX
\providecommand\bibfield[2]{#2}
\providecommand\bibinfo[2]{#2}
\providecommand\natexlab[1]{#1}
\providecommand\showeprint[2][]{arXiv:#2}
\makeatletter
\@ifundefined{NAT@parse@date}{}{\let\NAT@parse@date@orig\NAT@parse@date}
\@ifundefined{NAT@parse@date}{}{\def\NAT@parse@date#1#2#3#4#5#6@@{\NAT@parse@date@orig#1#2#3#4#5#6@@\def\NAT@tempyear{0000}\def\NAT@tempexlab{{?}}\ifx\NAT@year\NAT@tempyear\ifx\NAT@exlab\NAT@tempexlab\def\NAT@date{[n.\,d.]}\else\edef\NAT@date{[n.\,d.]\NAT@exlab}\fi\fi}}
\makeatother

\bibitem[Bock(2024)]%
        {wearweb}
\bibfield{author}{\bibinfo{person}{Marius Bock}.}
  \bibinfo{year}{2024}\natexlab{}.
\newblock \bibinfo{title}{{WEAR} project page}.
\newblock \bibinfo{howpublished}{\url{https://mariusbock.github.io/wear/}}.
\newblock


\bibitem[Bock(2026)]%
        {wearchallenge}
\bibfield{author}{\bibinfo{person}{Marius Bock}.}
  \bibinfo{year}{2026}\natexlab{}.
\newblock \bibinfo{title}{3rd {WEAR} Dataset Challenge}.
\newblock \bibinfo{howpublished}{Challenge web page.
  \url{https://mariusbock.github.io/wear/challenge.html}}.
\newblock


\bibitem[Bock et~al\mbox{.}(2024)]%
        {bock2024wear}
\bibfield{author}{\bibinfo{person}{Marius Bock}, \bibinfo{person}{Hilde
  Kuehne}, \bibinfo{person}{Kristof Van~Laerhoven}, {and}
  \bibinfo{person}{Michael Moeller}.} \bibinfo{year}{2024}\natexlab{}.
\newblock \showarticletitle{{WEAR}: An Outdoor Sports Dataset for Wearable and
  Egocentric Activity Recognition}.
\newblock \bibinfo{journal}{\emph{Proceedings of the ACM on Interactive,
  Mobile, Wearable and Ubiquitous Technologies}} \bibinfo{volume}{8},
  \bibinfo{number}{4} (\bibinfo{year}{2024}), \bibinfo{pages}{1--21}.
\newblock
\href{https://doi.org/10.1145/3699776}{doi:\nolinkurl{10.1145/3699776}}


\bibitem[goodpjw2008(2026)]%
        {goodpjw2026notebook}
\bibfield{author}{\bibinfo{person}{goodpjw2008}.}
  \bibinfo{year}{2026}\natexlab{}.
\newblock \bibinfo{title}{Learned Links + Counts}.
\newblock \bibinfo{howpublished}{Kaggle notebook, 3rd {WEAR} Dataset Challenge
  @ {HASCA} 2026.
  \url{https://www.kaggle.com/competitions/3rd-wear-dataset-challenge-hasca-2026/code}}.
\newblock


\bibitem[habakan(2026)]%
        {habakan2026post}
\bibfield{author}{\bibinfo{person}{habakan}.} \bibinfo{year}{2026}\natexlab{}.
\newblock \bibinfo{title}{What Egocentric Video Embeddings Encode, and How to
  Combine Them with Inertial Sensors}.
\newblock \bibinfo{howpublished}{Discussion post and notebook, 3rd {WEAR}
  Dataset Challenge @ {HASCA} 2026, Kaggle.
  \url{https://www.kaggle.com/code/kansukehabano/what-egocentric-video-embeddings-encode}.
  Code: \url{https://github.com/habakan/what-egocentric-embeddings-encode}}.
\newblock


\bibitem[{Kaggle}(2026)]%
        {kaggle2026wear}
\bibfield{author}{\bibinfo{person}{{Kaggle}}.} \bibinfo{year}{2026}\natexlab{}.
\newblock \bibinfo{title}{3rd {WEAR} Dataset Challenge @ {HASCA} 2026}.
\newblock \bibinfo{howpublished}{Kaggle competition.
  \url{https://www.kaggle.com/competitions/3rd-wear-dataset-challenge-hasca-2026}}.
\newblock


\bibitem[Kr{\"a}henb{\"u}hl and Koltun(2011)]%
        {krahenbuhl2011densecrf}
\bibfield{author}{\bibinfo{person}{Philipp Kr{\"a}henb{\"u}hl} {and}
  \bibinfo{person}{Vladlen Koltun}.} \bibinfo{year}{2011}\natexlab{}.
\newblock \showarticletitle{Efficient Inference in Fully Connected {CRFs} with
  {G}aussian Edge Potentials}. In \bibinfo{booktitle}{\emph{Advances in Neural
  Information Processing Systems}}, Vol.~\bibinfo{volume}{24}.
  \bibinfo{pages}{109--117}.
\newblock


\bibitem[Kuhn(1955)]%
        {kuhn1955hungarian}
\bibfield{author}{\bibinfo{person}{Harold~W. Kuhn}.}
  \bibinfo{year}{1955}\natexlab{}.
\newblock \showarticletitle{The {H}ungarian Method for the Assignment Problem}.
\newblock \bibinfo{journal}{\emph{Naval Research Logistics Quarterly}}
  \bibinfo{volume}{2}, \bibinfo{number}{1--2} (\bibinfo{year}{1955}),
  \bibinfo{pages}{83--97}.
\newblock
\href{https://doi.org/10.1002/nav.3800020109}{doi:\nolinkurl{10.1002/nav.3800020109}}


\bibitem[Mugoya(2026)]%
        {mugoya2026post}
\bibfield{author}{\bibinfo{person}{Peter Mugoya}.}
  \bibinfo{year}{2026}\natexlab{}.
\newblock \bibinfo{title}{Video Groups the Seconds, {IMU} Names Them: What
  Carries the Score}.
\newblock \bibinfo{howpublished}{Discussion post, 3rd {WEAR} Dataset Challenge
  @ {HASCA} 2026, Kaggle.
  \url{https://www.kaggle.com/competitions/3rd-wear-dataset-challenge-hasca-2026/discussion}}.
\newblock


\bibitem[Pedregosa et~al\mbox{.}(2011)]%
        {pedregosa2011sklearn}
\bibfield{author}{\bibinfo{person}{Fabian Pedregosa}, \bibinfo{person}{Ga{\"e}l
  Varoquaux}, \bibinfo{person}{Alexandre Gramfort}, \bibinfo{person}{Vincent
  Michel}, \bibinfo{person}{Bertrand Thirion}, \bibinfo{person}{Olivier
  Grisel}, \bibinfo{person}{Mathieu Blondel}, \bibinfo{person}{Peter
  Prettenhofer}, \bibinfo{person}{Ron Weiss}, \bibinfo{person}{Vincent
  Dubourg}, \bibinfo{person}{Jake Vanderplas}, \bibinfo{person}{Alexandre
  Passos}, \bibinfo{person}{David Cournapeau}, \bibinfo{person}{Matthieu
  Brucher}, \bibinfo{person}{Matthieu Perrot}, {and}
  \bibinfo{person}{{\'E}douard Duchesnay}.} \bibinfo{year}{2011}\natexlab{}.
\newblock \showarticletitle{Scikit-learn: Machine Learning in {P}ython}.
\newblock \bibinfo{journal}{\emph{Journal of Machine Learning Research}}
  \bibinfo{volume}{12} (\bibinfo{year}{2011}), \bibinfo{pages}{2825--2830}.
\newblock


\bibitem[Shi et~al\mbox{.}(2023)]%
        {shi2023tridet}
\bibfield{author}{\bibinfo{person}{Dingfeng Shi}, \bibinfo{person}{Yujie
  Zhong}, \bibinfo{person}{Qiong Cao}, \bibinfo{person}{Lin Ma},
  \bibinfo{person}{Jia Li}, {and} \bibinfo{person}{Dacheng Tao}.}
  \bibinfo{year}{2023}\natexlab{}.
\newblock \showarticletitle{{TriDet}: Temporal Action Detection with Relative
  Boundary Modeling}. In \bibinfo{booktitle}{\emph{IEEE/CVF Conference on
  Computer Vision and Pattern Recognition (CVPR)}}.
  \bibinfo{pages}{18857--18866}.
\newblock
\href{https://doi.org/10.1109/CVPR52729.2023.01808}{doi:\nolinkurl{10.1109/CVPR52729.2023.01808}}


\bibitem[Wang et~al\mbox{.}(2023)]%
        {wang2023videomaev2}
\bibfield{author}{\bibinfo{person}{Limin Wang}, \bibinfo{person}{Bingkun
  Huang}, \bibinfo{person}{Zhiyu Zhao}, \bibinfo{person}{Zhan Tong},
  \bibinfo{person}{Yinan He}, \bibinfo{person}{Yi Wang}, \bibinfo{person}{Yali
  Wang}, {and} \bibinfo{person}{Yu Qiao}.} \bibinfo{year}{2023}\natexlab{}.
\newblock \showarticletitle{{VideoMAE V2}: Scaling Video Masked Autoencoders
  with Dual Masking}. In \bibinfo{booktitle}{\emph{IEEE/CVF Conference on
  Computer Vision and Pattern Recognition (CVPR)}}.
  \bibinfo{pages}{14549--14560}.
\newblock


\bibitem[Zhang et~al\mbox{.}(2022)]%
        {zhang2022actionformer}
\bibfield{author}{\bibinfo{person}{Chen-Lin Zhang}, \bibinfo{person}{Jianxin
  Wu}, {and} \bibinfo{person}{Yin Li}.} \bibinfo{year}{2022}\natexlab{}.
\newblock \showarticletitle{{ActionFormer}: Localizing Moments of Actions with
  Transformers}. In \bibinfo{booktitle}{\emph{European Conference on Computer
  Vision (ECCV)}} \emph{(\bibinfo{series}{Lecture Notes in Computer Science})}.
  \bibinfo{publisher}{Springer}, \bibinfo{pages}{492--510}.
\newblock
\href{https://doi.org/10.1007/978-3-031-19772-7_29}{doi:\nolinkurl{10.1007/978-3-031-19772-7_29}}


\bibitem[Zhou et~al\mbox{.}(2004)]%
        {zhou2004consistency}
\bibfield{author}{\bibinfo{person}{Dengyong Zhou}, \bibinfo{person}{Olivier
  Bousquet}, \bibinfo{person}{Thomas~Navin Lal}, \bibinfo{person}{Jason
  Weston}, {and} \bibinfo{person}{Bernhard Sch{\"o}lkopf}.}
  \bibinfo{year}{2004}\natexlab{}.
\newblock \showarticletitle{Learning with Local and Global Consistency}. In
  \bibinfo{booktitle}{\emph{Advances in Neural Information Processing
  Systems}}, Vol.~\bibinfo{volume}{16}. \bibinfo{pages}{321--328}.
\newblock


\end{thebibliography}

\onecolumn
\appendix

\section{Implementation details}
\label{app:impl}

Table~\ref{tab:impl} lists the settings of every trained part of the system. The decoder and the graphs are described in Sections~\ref{sec:graph} and~\ref{sec:decode}; their code is in the released repository.

\begin{table}[!ht]
\caption{Settings of the trained parts. OOF: out-of-fold window macro F1 (all participants, scored tiles) or successor top-1. Test: how test predictions are formed. All parts are used in the final submission except the last two rows.}
\label{tab:impl}
\footnotesize
\setlength{\tabcolsep}{3pt}
\begin{tabular}{p{2.6cm}p{3.6cm}p{6.6cm}p{2.3cm}p{1.3cm}}
\toprule
Part & Input & Model and training & Test & OOF \\
\midrule
Video MLP & pooled features of the compressed embeddings (mean, std, mean of first three and of last three frames, mean $|\Delta|$), standardised per participant & MLP 1120--512--256--19, dropout 0.3 (input 0.1), Adam, learning rate $10^{-3}$, weight decay $10^{-4}$, batch 256, 15 epochs, label smoothing 0.05, class weights & refit on all, 2 seeds & 0.543 \\
Video transformer & 15 frames, standardised per participant & 2 pre-norm transformer layers (width 256, 4 heads, feed-forward 512, dropout 0.2) plus a mean/std branch; AdamW, learning rate $5\cdot10^{-4}$, weight decay 0.05, one-cycle schedule, batch 512, 12 epochs, label smoothing 0.05, class weights, 2 seeds & mean of fold models & 0.578 \\
IMU MLP & 179 features (Section~\ref{sec:window}) and a copy standardised per participant and limb, limb one-hot; all four limbs of each tile & MLP 362--512--256--19, dropout 0.3, Adam, learning rate $10^{-3}$, batch 512, 12 epochs, class weights & refit on all, 2 seeds & 0.578 \\
IMU gradient boosting & as IMU MLP, limb as categorical feature & scikit-learn HistGradientBoosting: 250 iterations, learning rate 0.08, 31 leaves, min.\ 40 samples per leaf, L2 1.0, balanced classes, random half of the rows & refit on all & 0.614 \\
IMU 1D CNN & 3 axes, magnitude, standardised axes (7 channels) & 6 convolution layers (64 to 256 channels, batch norm, GELU, two max-pools), mean and max pooling, limb embedding of size 32; AdamW, learning rate $2\cdot10^{-3}$, weight decay 0.01, 12 epochs of 300k windows, batch 1024, rotations of up to $10^\circ$, 2 seeds & mean of fold models & 0.557 \\
Stage-1 fusion $p^{(1)}$ & log probabilities of the five models above & weighted mean with weights 0.2, 0.2, 0.2, 0.3, 0.1 (order as above) & same & 0.772 \\
Joint network 1 & 15 frames and the IMU window & video transformer branch and IMU CNN branch as above, joint head with auxiliary video and IMU heads; tiles and windows shifted by 0.5\,s; 15 epochs, 2 seeds & mean of fold models & 0.673 \\
Joint network 2 & as joint network 1 & same network; windows cut at random sub-second offsets, rotations and scaling of the IMU, feature noise; AdamW, learning rate $5\cdot10^{-4}$, weight decay 0.05, one-cycle schedule, batch 512, 30 epochs, 3 seeds & mean of fold models & 0.675 \\
Contrastive end-to-start embeddings & end and start descriptors of the frames (mean of three edge frames, edge frame, edge difference, window mean) & two linear towers per model, InfoNCE with temperature 0.1, 3{,}000 steps; one on the compressed embeddings, trained in each fold on CPU (output 256), one on the full embeddings on Kaggle (output 128, 2 seeds) & similarities of out-of-fold models; for the test, models trained on all & -- \\
First-pass successor model & pair features: video tail--head and predicted-frame similarities with their row and column ranks, IMU extrapolation distances, limb ids, stage-1 class similarity, contrastive end-to-start similarities & HistGradientBoosting: 300 iterations, learning rate 0.08, 63 leaves, min.\ 100 samples per leaf; candidates: 30 video, 15 predicted-frame, 12 IMU per anchor & refit on all & top-1 0.534 \\
Successor rescorer (final link model) & Section~\ref{sec:succ}; with limb ids and both bi-encoder similarities & same booster; all positive pairs and a random 20\% of the negative pairs & refit on all & top-1 0.641 \\
Video bi-encoder & 15 frames & end and start towers (transformer), embedding 128, InfoNCE with temperature 0.1, 1{,}200 steps, three runs of 3 seeds & similarities averaged over models & -- \\
IMU bi-encoder & 1\,s window of one limb, 8 edge samples given explicitly & end and start towers (1D CNN), embedding 32, InfoNCE with temperature 0.1, 2{,}000 steps, 3 seeds & similarities averaged over models & -- \\
Sequence model & 17 windows along the best successor and predecessor candidates & video projection, IMU CNN, limb embedding and link score, 2 transformer layers (width 256, 4 heads); 12 epochs, one-cycle schedule with maximum learning rate $5\cdot10^{-4}$, 2 seeds & mean of fold models & 0.787 \\
Stage 2 (chains A, B, C) & 2{,}062 (A) or 2{,}081 (B, C) features, Section~\ref{sec:stacking} & MLP with hidden layers 512 and 256, dropout 0.3 (input 0.1), Adam, learning rate $10^{-3}$, weight decay $10^{-4}$, batch 256, 12 epochs, class weights & mean of fold models & 0.877, 0.882, 0.888 \\
Decoder inside stacking & mean of stage-2 and stage-1 log probabilities & label propagation ($\alpha=0.8$, 30 steps) on $A_s$ with $\lambda=10$; Ward clustering into $\max(20, n/25)$ groups connected in the graph; assignment of groups to classes with bounds 70 and 125 by a Lagrangian heuristic (300 steps) & same & -- \\
Stage 3 (chains A, B, C) & 461 (A) or 480 (B, C) features, Section~\ref{sec:stacking} & HistGradientBoosting: 150 iterations, learning rate 0.1, 31 leaves, min.\ 50 samples per leaf, L2 1.0, balanced classes & refit on all & 0.918, 0.918, 0.923 \\
Hungarian feature & membership in a one-to-one assignment of the score matrix & added to the rescorer (top-1 0.586); tested, not in the final submission & -- & -- \\
Seam features & Section~\ref{sec:seam} & added to the final link model (top-1 0.667); tested, not in the final submission & -- & -- \\
\bottomrule
\end{tabular}
\end{table}

\section{Per-class and per-participant results}
\label{app:classes}

Table~\ref{tab:classes} gives the per-class scores of the final system, the window counts per activity in the training data, the count errors with and without the count constraint, and the effect of the constraint on each class. Table~\ref{tab:participants} gives the macro F1 of each participant.

\begin{table}[!ht]
\caption{Per-class results of the final system in simulation (all participants, scored tiles). Counts: windows per activity in the 17 single-performance training participants (10th percentile, median, 90th percentile). Count error: mean absolute difference between the predicted and the true number of windows of the activity over the same participants, without and with the count constraint. Gained / lost: windows of the class that became correct and that became wrong when the constraint was added. Missed to null: share of the missed windows of the class that were predicted as null.}
\label{tab:classes}
\footnotesize
\begin{tabular}{lrcccccccc}
\toprule
Class & Windows & Precision & Recall & F1 & Counts q10 / q50 / q90 & Count error without / with & Gained & Lost & Missed to null \\
\midrule
Null & 26{,}030 & 0.920 & 0.956 & 0.938 & -- & -- & 450 & 82 & -- \\
Jogging & 2{,}503 & 0.961 & 0.965 & 0.963 & 81 / 96 / 107 & 3.9 / 2.7 & 7 & 18 & 0.55 \\
Jogging (rotating arms) & 2{,}192 & 0.953 & 0.934 & 0.943 & 86 / 93 / 101 & 4.6 / 3.6 & 50 & 5 & 0.24 \\
Jogging (skipping) & 2{,}275 & 0.964 & 0.949 & 0.957 & 75 / 91 / 100 & 2.8 / 1.5 & 7 & 21 & 0.40 \\
Jogging (sidesteps) & 2{,}468 & 0.988 & 0.970 & 0.979 & 82 / 92 / 103 & 2.6 / 2.4 & 5 & 20 & 0.77 \\
Jogging (butt-kicks) & 2{,}153 & 0.951 & 0.941 & 0.946 & 79 / 91 / 96 & 4.1 / 3.1 & 16 & 2 & 0.73 \\
Stretching (triceps) & 2{,}327 & 0.988 & 0.969 & 0.979 & 87 / 95 / 103 & 2.1 / 3.9 & 1 & 30 & 1.00 \\
Stretching (lunging) & 2{,}441 & 0.985 & 0.928 & 0.955 & 89 / 98 / 114 & 8.5 / 7.1 & 32 & 44 & 0.99 \\
Stretching (shoulders) & 2{,}257 & 0.951 & 0.907 & 0.928 & 88 / 94 / 98 & 5.8 / 4.2 & 4 & 27 & 0.97 \\
Stretching (hamstrings) & 2{,}300 & 0.959 & 0.916 & 0.937 & 87 / 95 / 106 & 9.8 / 7.0 & 34 & 35 & 1.00 \\
Stretching (lumbar rotation) & 2{,}322 & 0.959 & 0.962 & 0.960 & 77 / 95 / 111 & 7.4 / 6.7 & 1 & 18 & 0.98 \\
Push-ups & 2{,}166 & 0.883 & 0.911 & 0.897 & 78 / 90 / 96 & 19.5 / 6.4 & 79 & 51 & 0.35 \\
Push-ups (complex) & 2{,}315 & 0.909 & 0.902 & 0.906 & 83 / 97 / 109 & 18.9 / 7.9 & 188 & 45 & 0.27 \\
Sit-ups & 2{,}393 & 0.946 & 0.878 & 0.910 & 84 / 92 / 104 & 15.5 / 9.4 & 135 & 21 & 0.73 \\
Sit-ups (complex) & 2{,}411 & 0.937 & 0.908 & 0.922 & 94 / 102 / 105 & 14.9 / 9.8 & 12 & 106 & 0.74 \\
Burpees & 2{,}333 & 0.946 & 0.930 & 0.938 & 87 / 93 / 106 & 6.9 / 5.4 & 32 & 22 & 0.96 \\
Lunges & 2{,}365 & 0.942 & 0.899 & 0.920 & 84 / 95 / 108 & 5.2 / 5.0 & 55 & 27 & 0.58 \\
Lunges (complex) & 2{,}435 & 0.946 & 0.907 & 0.926 & 84 / 96 / 109 & 5.8 / 4.8 & 32 & 26 & 0.72 \\
Bench-dips & 2{,}109 & 0.931 & 0.907 & 0.919 & 77 / 87 / 95 & 13.0 / 6.0 & 86 & 7 & 1.00 \\
\bottomrule
\end{tabular}
\end{table}

\begin{table}[!ht]
\caption{Macro F1 of the final system for each training participant in simulation (scored tiles, over the classes that occur in the participant's labels or predictions). Participants marked with an asterisk performed some or all activities twice.}
\label{tab:participants}
\footnotesize
\setlength{\tabcolsep}{3.2pt}
\begin{tabular}{lccccccccccc}
\toprule
Participant & 0* & 1 & 2* & 3* & 4 & 5 & 6 & 7 & 8 & 9 & 10* \\
Macro F1 & 0.915 & 0.905 & 0.964 & 0.947 & 0.796 & 0.947 & 0.962 & 0.882 & 0.971 & 0.947 & 0.854 \\
\midrule
Participant & 11 & 12 & 13 & 14* & 15 & 16 & 17 & 18 & 19 & 20 & 21 \\
Macro F1 & 0.972 & 0.946 & 0.950 & 0.932 & 0.960 & 0.962 & 0.974 & 0.943 & 0.969 & 0.972 & 0.974 \\
\bottomrule
\end{tabular}
\end{table}

\end{document}